\documentclass[letterpaper]{article} 
\usepackage{aaai2026}  
\usepackage{times}  
\usepackage{helvet}  
\usepackage{courier}  
\usepackage[hyphens]{url}  
\usepackage{graphicx} 
\usepackage{natbib}  
\usepackage{caption} 
\usepackage{algorithm}
\usepackage{algpseudocode}
\usepackage{amsmath}
\usepackage{multirow}
\usepackage{booktabs}
\usepackage{xcolor}
\usepackage{listings}
\usepackage{newfloat}
\usepackage{listings}
\DeclareCaptionStyle{ruled}{labelfont=normalfont,labelsep=colon,strut=off} 
\floatstyle{ruled}
\newfloat{listing}{tb}{lst}{}
\floatname{listing}{Listing}
\title{Extracting Events Like Code: A Multi-Agent Programming Framework for Zero-Shot Event Extraction}
\author {
    Quanjiang Guo\textsuperscript{\rm 1},
    Sijie Wang\textsuperscript{\rm 2},
    Jinchuan Zhang\textsuperscript{\rm 1},
    Ben Zhang\textsuperscript{\rm 1},
    Zhao Kang\textsuperscript{\rm 1},
    Ling Tian\textsuperscript{\rm 1},
    Ke Yan\textsuperscript{\rm 1}\thanks{Corresponding author}
}
\affiliations {
    \textsuperscript{\rm 1}University of Electronic Science and Technology of China, China\\
    \textsuperscript{\rm 2}Nanyang Technological University, Singapore\\
    guochance1999@163.com, wang1679@e.ntu.edu.sg, \{jinchuanz, zhangben\}@std.uestc.edu.cn, \\
    \{zkang, lingtian, kyan\}@uestc.edu.cn
}

\usepackage{bibentry}

\begin{document}

\maketitle

\begin{abstract}
Zero-shot event extraction (ZSEE) remains a significant challenge for large language models (LLMs) due to the need for complex reasoning and domain-specific understanding. Direct prompting often yields incomplete or structurally invalid outputs—such as misclassified triggers, missing arguments, and schema violations. To address these limitations, we present Agent-Event-Coder (AEC), a novel multi-agent framework that treats event extraction like software engineering: as a structured, iterative code-generation process. AEC decomposes ZSEE into specialized subtasks—retrieval, planning, coding, and verification—each handled by a dedicated LLM agent. Event schemas are represented as executable class definitions, enabling deterministic validation and precise feedback via a verification agent. This programming-inspired approach allows for systematic disambiguation and schema enforcement through iterative refinement. By leveraging collaborative agent workflows, AEC enables LLMs to produce precise, complete, and schema-consistent extractions in zero-shot settings. Experiments across five diverse domains and six LLMs demonstrate that AEC consistently outperforms prior zero-shot baselines, showcasing the power of treating event extraction like code generation\footnote{The code and data are released on \url{https://github.com/UESTC-GQJ/Agent-Event-Coder}.}.
\end{abstract}

\section{Introduction}
\begin{figure}[htbp]
	\centerline{\includegraphics[width=1\columnwidth, scale=0.8]{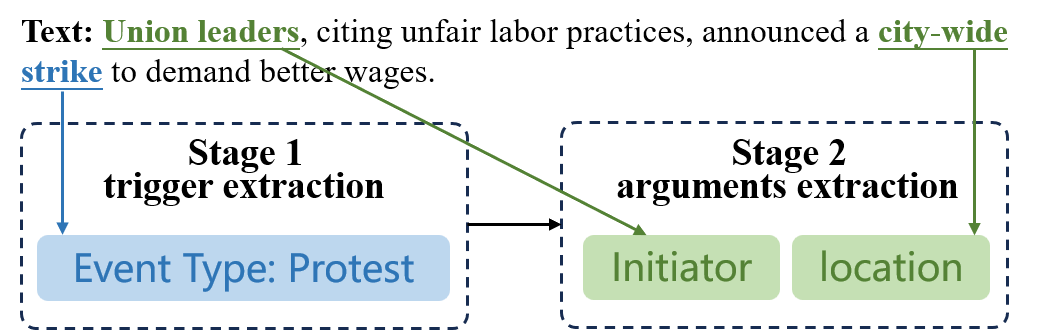}}
	\caption{An illustrative example of the event extraction task. The blue box denotes the event type, while the green boxes represent the argument roles. Underlined words indicate the event trigger or the corresponding event arguments.
}
	\label{fig:intro}
\end{figure}

Event extraction (EE) aims to identify event triggers and their associated arguments from unstructured text~\cite{xu2024large}, we provide an illustration of the task in Figure~\ref{fig:intro}. As a structured prediction task, it plays a vital role in applications such as knowledge base population, information retrieval, and question answering. Traditional EE methods rely on supervised learning and require labeled examples for each event type. However, the growing diversity of event types and the high cost of annotation make it impractical to collect training data for all possible events.

\begin{figure}[ht]
	\centerline{\includegraphics[width=1\columnwidth, scale=0.8]{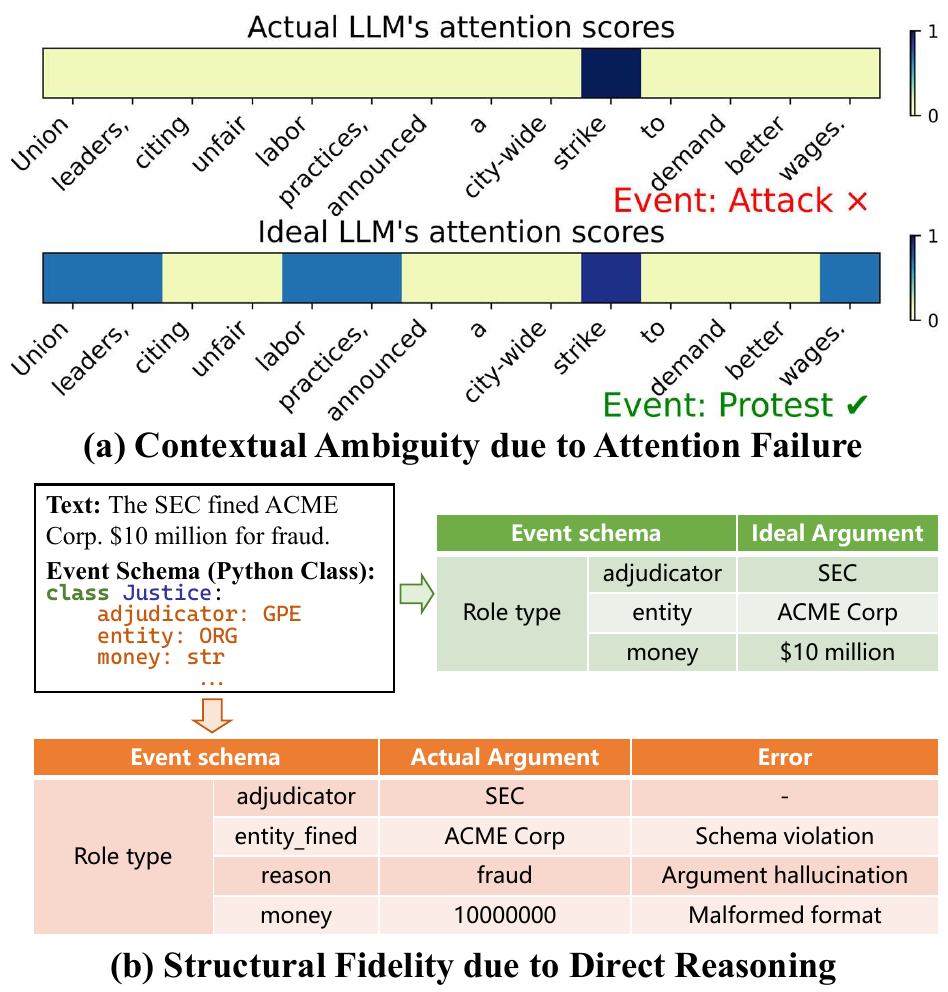}}
	\caption{(a) Conceptual illustration of attention failure. The ideal model effectively leverages contextual information to correctly interpret the trigger word \textit{“strike”} as an instance of the \textit{Protest} event type. In contrast, direct zero-shot prompting of LLMs tends to over-rely on the trigger word itself, often leading to misclassification.
(b) Illustration of extraction errors caused by insufficient structural fidelity. Outputs generated by direct zero-shot prompting of LLMs may violate the target event schema by: (1) including a non-existent argument role, (2) hallucinating an undefined argument, or (3) producing an argument with an incorrect data type.}
	\label{fig:intro1}
\end{figure}

Zero‑shot event extraction (ZSEE) seeks to address the limitations of supervised event extraction by enabling models to identify event types that have not been observed during training, using only the event type’s name or natural language definition. Rather than relying on annotated examples, the model must leverage these textual definitions or ontological descriptions of event types and roles to guide prediction. Although ZSEE offers significant potential for scalable event extraction, it remains highly challenging~\cite{chen2024large, cai-etal-2024-improving-event}, and existing approaches frequently fall short due to two key issues. (i) \textit{Contextual ambiguity}: candidate event trigger words are often polysemous, and their correct interpretation depends on subtle contextual cues. For example, the word “strike” may refer to either a labor protest or a physical attack depending on the surrounding context. In the absence of training examples, capturing such nuances is difficult. As illustrated in Figure~\ref{fig:intro1}(a), large language models (LLMs) may misinterpret the trigger or fail to exploit contextual clues that indicate the correct event type and its arguments. (ii) \textit{Structural fidelity}: event extraction is a structured prediction task that requires outputs to conform to a predefined schema, such as a JSON object or database entry. While LLMs can be prompted to produce structured outputs, they often fail to strictly follow schema constraints, particularly without fine‑tuning~\cite{liu-etal-2024-unleashing-power, ijcai2025p897}. This can result in malformed or incomplete event records that disrupt downstream processing. As shown in Figure~\ref{fig:intro1}(b), direct zero‑shot prompting of LLMs frequently leads to misidentified triggers or invalid output structures, highlighting the need for a more guided and robust approach to zero‑shot event extraction.

To address these limitations, we introduce \textbf{Agent‑Event‑Coder (AEC)}, a novel framework that reconceptualizes zero‑shot event extraction as a collaborative and verifiable code‑generation process. Instead of prompting a single LLM to directly produce a structured output, AEC decomposes the extraction pipeline into four specialized agents—Retrieval, Planning, Coding, and Verification—each responsible for a distinct subtask. AEC is built upon two core principles. \textbf{(i) Multi‑agent decomposition:} the overall task is divided into interpretable reasoning stages. For instance, the Planning Agent generates trigger–type hypotheses accompanied by explanatory rationales, while the Coding Agent converts the highest‑confidence hypothesis into executable Python code that instantiates a schema‑compliant event class. \textbf{(ii) Schema‑as‑code verification:} event schemas are represented as executable \texttt{Python} classes, enabling deterministic structural validation at runtime. A dedicated Verification Agent evaluates the generated code for semantic compatibility, type correctness, and structural validity. When validation fails, a dual‑loop refinement procedure is initiated, iteratively patching the code based on compiler‑like diagnostic feedback and, if necessary, exploring lower‑confidence hypotheses. By combining step‑wise reasoning with deterministic schema validation, this programming‑inspired architecture allows AEC to systematically resolve trigger ambiguity and enforce structural fidelity in zero‑shot settings, producing precise, complete, and schema‑compliant event extractions without requiring annotated training examples.

In summary, we make the following contributions: 
\begin{itemize}

    \item We are the \textbf{first} to reformulate ZSEE as a multi-agent code generation task, providing a new paradigm that unifies schema constraints, iterative planning, and code-based validation for structured event extraction.
    
    \item We introduce a new multi-agent workflow \textbf{AEC}, where specialized agents work together to retrieve knowledge, design extraction plans, and generate structured event representations.
    
    \item We design a schema-as-code verification loop, in which a dedicated verification agent applies deterministic programming language rules to constraint outputs and provide precise feedback for iterative refinement.
    
    \item Through comprehensive evaluations across five diverse domains and six LLMs, we demonstrate the robustness, generalizability, and effectiveness of AEC as a state-of-the-art ZSEE framework.

\end{itemize}

\section{Related Work}
\subsection{ZSEE with Prompting}
Recent works have explored using LLMs for information extraction tasks by formulating extraction as a prompting or question-answering problem. In event extraction, approaches such as ChatIE~\cite{wei2023chatie} engage in structured dialogues with ChatGPT to iteratively refine event outputs, while others, such as CODE4STRUCT~\cite{wang2023code4struct} and Code4UIE~\cite{guo2024retrieval}, represent events and schemas as code or templates to leverage the reasoning capabilities of LLMs. These methods enable models to perform zero-shot or few-shot extraction by providing task descriptions or examples to the model. Additional studies have incorporated event definitions or constraints into prompts to guide the model—for example, by using positive and negative instructions about the event type and trigger~\cite{srivastava2025instruction}. However, purely prompt-based single-agent strategies struggle with complex structured tasks~\cite{guo2025baner}. Without explicit decomposition, an LLM may overlook subtle interactions between an event’s trigger and its arguments. Moreover, these methods are highly sensitive to prompt design and the choice of demonstration examples, which can easily mislead the model in zero-shot settings.

\begin{figure*}[ht]
	\centerline{\includegraphics[width=2.1\columnwidth]{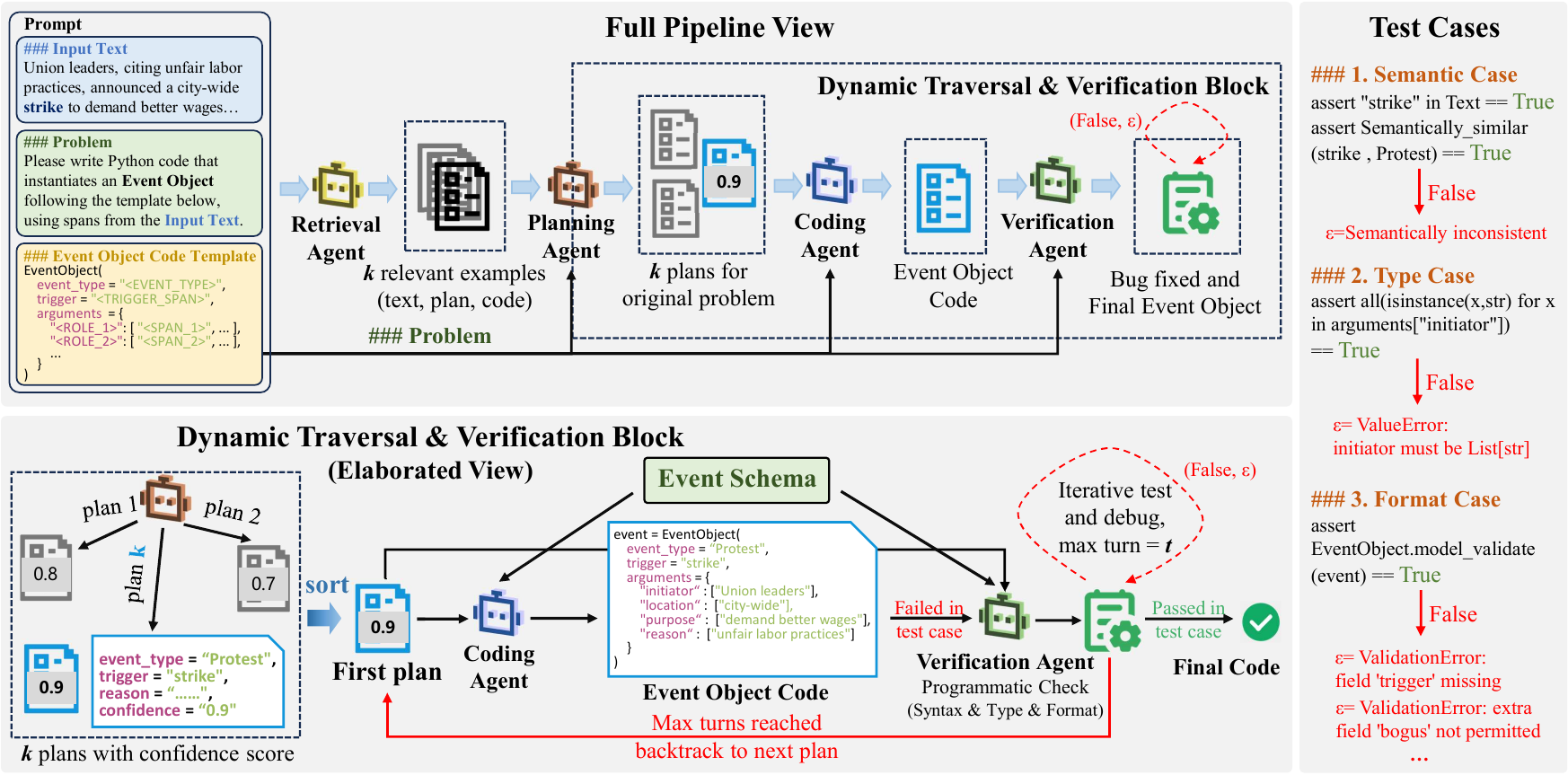}}
	\caption{Overview of the proposed AEC framework. The \textbf{Full Pipeline View} (top) illustrates four specialized agents—\textit{Retrieval}, \textit{Planning}, \textit{Coding}, and \textit{Verification}—collaborating to generate schema-compliant event objects from unstructured text. The \textit{Retrieval Agent} self-generates relevant exemplars to bridge the gap between schema definitions and textual context. The \textit{Planning Agent} produces $\textbf{\textit{k}}$ trigger–type hypotheses, each with a confidence score and explanatory rationale. The \textit{Coding Agent} converts the highest-confidence hypothesis into executable Python code that instantiates a predefined event schema. The \textbf{Dynamic Traversal and Verification Block} (bottom) depicts the iterative refinement loop. The generated code is evaluated by the \textit{Verification Agent} through three deterministic \textbf{test cases}: \textit{semantic}, \textit{type}, and \textit{format} checks (right). If a test fails, the agent patches the code using compiler-like diagnostics. When refinement attempts for the current plan are exhausted, the system backtracks to the next-best hypothesis. This dual-loop architecture ensures that the final output satisfies both semantic correctness and structural fidelity—\textit{without requiring any labeled examples}.}

	\label{fig:methodology}
\end{figure*}
\subsection{Multi-Agent Collaboration for Information Extraction}
Multi-agent systems have been proposed to improve information extraction by having specialized agents or models cooperate or debate to reach better results. ~\cite{talebirad2023multi} explore frameworks in which multiple LLM agents collaborate (either cooperatively or adversarially) through iterative dialogues, showing that such interactions can refine outputs for complex reasoning tasks. In event extraction, ~\cite{wang-huang-2024-debate} introduce a debate-style optimization (DoA) in a few-shot setting, where two agents discuss and revise event predictions to reduce errors. For relation extraction, ~\cite{hou2024multiagent} propose a dual-agent approach (EPASS) that jointly models entity-pair extraction and supporting-evidence identification, demonstrating the benefit of agents focusing on different subtasks. ~\cite{lu2024triageagent} present TriageAgent, a heterogeneous multi-agent system for clinical IE, in which multiple LLM-based agents role-play with turn-taking, confidence scoring, and early stopping criteria to extract medical events more accurately. Most relevant to our work, ~\cite{wang2025cooperative} recently applied a cooperative multi-agent system to zero-shot named entity recognition (NER). Their framework (CMAS) uses four agents to handle entity-span detection, type-specific feature extraction, demonstration discrimination, and final prediction, yielding improved zero-shot NER performance by addressing context correlations and filtering prompt examples. 

AEC adopts this collaborative paradigm but adapts it to the unique challenges of event extraction, where predictions must account for both triggers and multiple arguments under strict schema constraints. Unlike existing multi‑agent IE frameworks, AEC introduces dedicated agents for trigger–type hypothesis generation, event coding, and schema‑level verification, thereby integrating step‑wise reasoning with deterministic output validation. To the best of our knowledge, AEC is the first framework to employ a multi‑agent, code‑generation‑based strategy for ZSEE, enabling both contextual disambiguation and structural fidelity through its division of labor and iterative refinement process.

\section{Methodology}

\noindent\textbf{Task Definition.} ZSEE takes as input an unstructured text span $T = {w_1,\dots,w_n}$ and an unseen event schema $S_e = \langle e, R_e\rangle$, where $e$ is the event type and $R_e={(r_j,\tau_j)}_{j=1}^m$ denotes a set of argument roles $r_j$ with their expected value types $\tau_j$. The objective is to generate a fully specified event instance:
\begin{equation}
    y = \langle e, z, A \rangle,
\end{equation}
where $z \in T$ is the predicted trigger span and $A={(r_j,a_j)}_{j=1}^m$ is the set of argument-role pairs with $a_j \subseteq T$ or $a_j=\emptyset$. No labeled examples for $e$ are available.

\subsection{Overall Architecture}

As illustrated in Figure ~\ref{fig:methodology}, AEC addresses ZSEE through a structured four‑agent pipeline, designed with two nested feedback loops that enable iterative reasoning and verification.

\noindent\textbf{Retrieval Agent.}~The retrieval agent $A_{\text{ret}}$ is responsible for \textit{self‑generating} a set of $\textbf{\textit{k}}$ high‑quality exemplar sentences tailored to the given event schema $S_{e}$:
\begin{equation}
      A_{\text{ret}}(S_{e})\;\rightarrow\;
      D_{\text{ex}}=\{\,s_{1},\dots,s_{k}\,\},
\end{equation}
Drawing inspiration from the \emph{analogical prompting} paradigm proposed by \citet{yasunaga_large_2024}, these exemplars serve as schema–textual “analogies” that align abstract constraints with concrete linguistic realizations. By embedding step‑by‑step guidance into the demonstration space, they help disambiguate polysemous triggers and ground the model’s reasoning in context, thereby reducing early commitment errors and improving planning agent performance.

\noindent\textbf{Planning Agent.}~The planning agent $A_{\text{plan}}$ examines the input text $T$ in the context of the retrieved exemplars $D_{\text{ex}}$ and, by leveraging both lexical and semantic cues, produces a ranked list of \emph{trigger–type} hypotheses accompanied by natural‑language rationales:
\begin{equation}
      A_{\text{plan}}(T,S_{e},D_{\text{ex}})\!\rightarrow\!
      P=\bigl\{\,\bigl((z_{i},e),\,\beta_{i},\,\rho_{i}\bigr)\bigr\}_{i=1}^{k},
\end{equation}

where each $z_{i}$ denotes a potential trigger in $T$, $\beta_{i}\!\in\![0,1]$ represents the model‑assigned confidence that $z_{i}$ evokes event type $e$, and $\rho_{i}$ provides a concise natural‑language explanation for why the pair $(z_{i},e)$ is plausible. These rationales $\rho_{i}$ are retained for subsequent error analysis and ablation studies, enabling a better understanding of the agent’s decision‑making process.

\begin{figure}[htbp]
	\centerline{\includegraphics[width=1\columnwidth, scale=0.8]{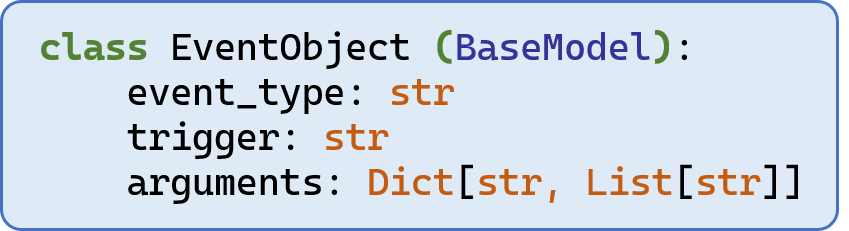}}
	\caption{Fixed output template used by all agents.}
	\label{fig:method}
\end{figure}

\noindent\textbf{Coding Agent.} Following ~\cite{srivastava2025instruction}, we compile every new schema
$S_{e}=\bigl\langle e,\{(r_{1},\tau_{1}),\dots,(r_{m},\tau_{m})\}\bigr\rangle$ into a Python
\texttt{BaseModel} whose constructor enforces role types~$\tau_{j}$.
Schema compliance therefore reduces to constructing a valid class instance,
which can be \emph{deterministically} verified at run time. ~$A_{\text{code}}$ converts the highest‑scoring hypothesis $((z^{\star},e),\beta^{\star},\rho^{\star})$ into executable Python that instantiates a Python template, as shown in Figure ~\ref{fig:method}.

\noindent\textbf{Verification Agent.}~The verification agent $A_{\text{verify}}$ evaluates the generated code object $C_{\text{obj}}$ by executing a comprehensive three‑stage test suite that checks semantic correctness, type validity, and structural integrity. Based on the outcome of these checks, the agent produces a binary verdict and, in case of failure, a diagnostic message $\varepsilon$ indicating the first failed test:
\begin{equation}
      A_{\text{verify}}(C_{\text{obj}})\;\rightarrow\;(V,\varepsilon),
      \qquad V\in\{\texttt{True},\texttt{False}\}.
\end{equation}
This design ensures that only code satisfying all verification criteria is accepted, while informative feedback is provided to guide subsequent error correction.

\subsection{Three‑Stage Verification}

The verification agent $A_{\text{verify}}$ executes a test suite consisting of three checks:

\begin{itemize}
\item \textbf{Semantic Check ${\cal T}_{1}$:} 
      Ensures that the predicted trigger $z^{\star}$ appears in the input text $T$ and is semantically compatible with the event type $e$, based on lexical matching and contextual similarity.
\item \textbf{Type Check ${\cal T}_{2}$:} 
      Verifies that each argument value conforms to the datatype specified in the schema $S_{e}$, while multiplicity constraints are enforced via Pydantic validation.
\item \textbf{Structural Check ${\cal T}_{3}$:} 
      Confirms that the generated code compiles successfully, contains exactly the fields \{\texttt{event\_type}, \texttt{trigger}, \texttt{arguments}\}, and produces a serializable event object.
\end{itemize}

The agent returns $V=\texttt{True}$ only if all three checks ${\cal T}_{1}\!\land{\cal T}_{2}\!\land{\cal T}_{3}$ succeed; otherwise, $\epsilon$ indicates the first failed test.

\subsection{Dual‑Loop Refinement Algorithm}

Let $\textbf{\textit{k}}$ be the number of hypotheses and $t$ the maximum patch attempts per hypothesis, the selection and verification procedure is shown in algorithm~\ref{alg:selection}.

\begin{algorithm}[htbp]
\caption{Dual‑Loop Refinement Algorithm}
\label{alg:selection}
\begin{algorithmic}[1]
\Require Candidate pool $P$
\Ensure Valid code object $C_{\text{obj}}$
\State Pick $((z^{\star},e),\beta^{\star},\rho^{\star})=\arg\max\limits_{P} \beta$ \Comment{Selection}
\While{$P \neq \emptyset$}
    \For{$j=1$ to $t$} \Comment{Inner Loop}
        \State Generate code $C_{\text{obj}}$ with $A_{\text{code}}$
        \State Get $(V,\epsilon) = A_{\text{verify}}(C_{\text{obj}})$
        \If{$V$ is true}
            \State \Return $C_{\text{obj}}$
        \Else
            \State Patch code using $\varepsilon$
        \EndIf
    \EndFor
    \State Remove current hypothesis from $P$
    \State Pick $((z^{\star},e),\beta^{\star},\rho^{\star})=\arg\max\limits_{P} \beta$
\EndWhile
\end{algorithmic}
\end{algorithm}

The algorithm explores $O($\textbf{\textit{k}}$t)$ candidate paths yet guarantees that the final output is both semantically correct and schema‑consistent, thereby realising reliable ZSEE without labelled examples.

\begin{table*}[ht]
\setlength{\tabcolsep}{0.9mm}
\renewcommand{\arraystretch}{1.2}
\centering
\begin{tabular}{cc|cc|cccc|cccc|cc|cccc}
\hline
\multirow{2}{*}{LLM}        & \multirow{2}{*}{Strategy} & \multicolumn{2}{c|}{FewEvent(100)} & \multicolumn{4}{c|}{ACE(33)}                                 & \multicolumn{4}{c|}{GENIA(9)}                                                                         & \multicolumn{2}{c|}{SPEED(7)} & \multicolumn{4}{c}{CASIE(5)}                                 \\
                            &                           & TI                 & TC             & TI            & TC            & AI            & AC            & TI            & TC            & AI                                & AC                                 & TI             & TC            & TI            & TC            & AI            & AC            \\ \hline
\multirow{6}{*}{Llama3-8B}  & DirectEE                  & 21.5               & 17.5           & 26.4          & 25.7          & -             & -             & 27.8          & 27.4          & -                                 & -                                  & 34.3           & 41.5          & 11.8          & 47.6          & -             & -             \\
                            & GuidelineEE               & 15.8               & 16.4           & 32.4          & 30.5          & 25.4          & 23.7          & 27.1          & 26.7          & \multicolumn{1}{l}{22.4}          & \multicolumn{1}{l|}{\textbf{21.8}} & 35.0           & 36.9          & 12.5          & 43.1          & \textbf{31.7} & 27.8          \\
                            & DecomposeEE               & 20.7               & 20.5           & 30.1          & 35.2          & 26.1          & 24.9          & 28.9          & 28.5          & \multicolumn{1}{l}{22.8}          & \multicolumn{1}{l|}{21.7}          & 31.2           & 38.4          & 10.5          & 50.7          & 27.0          & 25.1          \\
                            & CEDAR                     & 25.2               & 18.7           & 36.1          & 30.9          & -             & -             & 29.8          & 29.4          & -                                 & -                                  & 34.9           & 37.3          & 15.8          & 48.3          & -             & -             \\
                            & ChatIE                    & -                  & 24.8           & -             & 44.2          & 32.4          & 30.8          & -             & 23.8          & \multicolumn{1}{l}{21.7}          & \multicolumn{1}{l|}{20.3}          & -              & \textbf{42.9} & -             & 33.3          & 22.2          & 20.8          \\
                            & \textbf{AEC}              & \textbf{27.0}   & \textbf{27.6}  & \textbf{40.5} & \textbf{48.8} & \textbf{33.7} & \textbf{31.8} & \textbf{32.0} & 31.5          & \textbf{25.3}                     & \multicolumn{1}{l|}{21.5}          & \textbf{36.3}  & 41.8          & \textbf{16.5} & \textbf{55.7} & 30.7          & \textbf{28.5} \\ \hline
\multirow{6}{*}{Llama3-70B} & DirectEE                  & 32.1               & 30.3           & 50.7          & 46.9          & -             & -             & 37.3          & 45.3          & -                                 & -                                  & 44.7           & 48.9          & 13.5          & 62.4          & -             & -             \\
                            & GuidelineEE               & 29.8               & 32.1           & 46.2          & 51.4          & 30.7          & 27.5          & 31.5          & 34.8          & 27.5                              & 27.0                               & 41.5           & 44.1          & 12.3          & 48.3          & 36.2          & 31.6          \\
                            & DecomposeEE               & 35.3               & 33.5           & 45.8          & 49.3          & 30.4          & 27.9          & 39.2          & 44.8          & 28.3                              & 26.9                               & 40.0           & 42.8          & 17.5          & 60.7          & 33.5          & 31.9          \\
                            & CEDAR                     & 34.5               & 33.9           & 51.5          & 48.7          & -             & -             & 36.8          & 47.7          & -                                 & -                                  & \textbf{45.3}  & 49.8          & 16.7          & 54.3          & -             & -             \\
                            & ChatIE                    & -                  & \textbf{40.7}  & -             & 47.5          & 36.6          & 34.5          & -             & 34.2          & \multicolumn{1}{l}{27.9}          & \multicolumn{1}{l|}{26.1}          & -              & 50.5          & -             & 50.8          & 28.7          & 25.5          \\
                            & \textbf{AEC}              & \textbf{42.1}      & 40.5           & \textbf{57.0} & \textbf{54.6} & \textbf{38.4} & \textbf{34.7} & \textbf{39.4} & \textbf{48.1} & \multicolumn{1}{l}{\textbf{31.2}} & \multicolumn{1}{l|}{\textbf{30.1}} & 43.8           & \textbf{52.3} & \textbf{18.7} & \textbf{65.9} & \textbf{36.4} & \textbf{33.9} \\ \hline
\end{tabular}
\caption{Main results comparing the ZSEE performance of our proposed AEC with all other baselines for the Llama3-8B-Instruct and Llama3-70B-Instruct LLMs. bold = best performance. ($\cdot$) = number of distinct event types.}
\label{table:main_results}
\end{table*}

\section{Experiments}
\subsection{Experimental Setup}
\paragraph{Datasets.} We evaluate AEC on five widely used event extraction benchmarks covering diverse domains: FewEvent (General)~\cite{deng2020meta}, ACE 2005 (News)~\cite{doddington2004automatic}, GENIA (Biomedical), SPEED (Epidemiological), and CASIE (Cybersecurity). For datasets without argument annotations (FewEvent and SPEED), we report only Trigger Identification (TI) and Trigger Classification (TC). For ACE 2005, GENIA, and CASIE, we additionally evaluate Argument Identification (AI) and Argument Classification (AC).

To mitigate potential distributional biases, we follow the \textsc{TextEE} evaluation protocol~\cite{huang2024textee, parekh2025dicore} and uniformly sample 250 test instances from each dataset to form evaluation splits. For CASIE, due to its smaller size, we sample 50 test instances. Our experiments are conducted under a purely zero‑shot setting, i.e., no training data are used.

\paragraph{Baselines.} 
We benchmark AEC against five strong zero-shot event extraction baselines:

\begin{itemize}

\item \textbf{DirectEE}~\cite{gao2023exploring} prompts LLMs directly to extract structured events in a single inference step without intermediate reasoning or decomposition.

\item \textbf{CEDAR}~\cite{li2023glen} adopts a multi-stage detection framework explicitly designed for large-ontology event detection, involving hierarchical reasoning over event types and triggers.

\item \textbf{DecomposeEnrichEE}~\cite{shiri2024decompose} decomposes event extraction into event detection and argument extraction stages, utilizing dynamic, schema-aware retrieval augmentation to reduce hallucinations.

\item \textbf{GuidelineEE}~\cite{srivastava2025instruction} leverages annotation guidelines, converting event extraction into a structured Python code-generation task guided by textual schema descriptions.

\item \textbf{ChatIE}~\cite{wei2023chatie} transforms zero-shot IE into a conversational, multi-turn question-answering process, iteratively querying the LLM to progressively refine extraction outputs.

\end{itemize}

For fair comparison, all baselines were adapted to output structured, schema-conformant event objects. Moreover, we incorporate a unified Verification component into each baseline (if not already present) to ensure robustness and consistency in benchmarking performance.

\begin{table*}[ht]
\setlength{\tabcolsep}{0.85mm}
\renewcommand{\arraystretch}{1.3}
\centering
\begin{tabular}{ll|cc|cclc|cllc|cc|cclc}
\hline
\multirow{2}{*}{\textbf{LLM}} & \textbf{Prompt} & \multicolumn{2}{c|}{FewEvent(100)} & \multicolumn{4}{c|}{ACE(33)}                                                        & \multicolumn{4}{c|}{GENIA(9)}                                                             & \multicolumn{2}{c|}{SPEED(7)} & \multicolumn{4}{c}{CASIE(5)}                                 \\
                              & \textbf{Style}  & TI               & TC               & TI            & TC            & AI                       & AC                        & TI            & \multicolumn{1}{c}{TC} & AI                                & AC            & TI             & TC            & TI            & TC            & AI            & AC            \\ \hline
\multirow{3}{*}{Qwen2.5-14B}  & GuidelineEE     & 23.5             & 20.2             & 35.1          & 33.5          & \multicolumn{1}{c}{22.4} & 21.0                      & 27.0          & 25.8                   & \multicolumn{1}{c}{20.1}          & 19.2          & 32.5           & 36.4          & 15.2          & 47.5          & 26.8          & 27.3          \\
                              & DecomposeEE     & 25.7             & 23.9             & 37.6          & 38.8          & \multicolumn{1}{c}{25.1} & \textbf{23.2}             & 29.3          & 28.5                   & \multicolumn{1}{c}{23.2}          & 22.0          & 34.7           & 38.9          & 16.4          & 49.8          & 27.9          & 28.8          \\
                              & AEC             & \textbf{30.4}    & \textbf{28.1}    & \textbf{42.5} & \textbf{45.3} & \textbf{28.5}            & 22.7                      & \textbf{32.1} & \textbf{30.7}          & \multicolumn{1}{c}{\textbf{25.6}} & \textbf{24.8} & \textbf{38.6}  & \textbf{42.4} & \textbf{17.8} & \textbf{53.2} & \textbf{29.5} & \textbf{31.4} \\ \hline
\multirow{3}{*}{Qwen2.5-72B}  & GuidelineEE     & 34.8             & 32.2             & 47.6          & 50.2          & \multicolumn{1}{c}{30.8} & 29.4                      & 35.5          & 34.6                   & 28.0                              & 27.2          & 40.6           & 45.7          & 17.6          & 54.3          & 33.7          & 32.4          \\
                              & DecomposeEE     & 37.1             & 35.8             & 49.9          & 53.7          & 31.6                     & 32.7                      & 38.4          & 36.9                   & 29.7                              & 28.6          & 43.1           & 47.9          & 18.2          & 57.5          & 34.8          & 33.7          \\
                              & AEC             & \textbf{39.8}    & \textbf{38.4}    & \textbf{54.3} & \textbf{58.0} & \textbf{36.4}            & \textbf{34.9}             & \textbf{40.7} & \textbf{39.4}          & \textbf{31.9}                     & \textbf{30.1} & \textbf{47.6}  & \textbf{52.5} & \textbf{20.5} & \textbf{60.8} & \textbf{37.5} & \textbf{35.4} \\ \hline
\multirow{3}{*}{GPT3.5-turbo} & GuidelineEE     & 23.2             & 20.9             & 35.1          & 37.9          & 20.6                     & 17.4                      & 26.7          & 25.5                   & 19.1                              & 18.3          & 31.3           & 36.5          & 18.3          & 56.7          & \textbf{31.5} & \textbf{30.8}          \\
                              & DecomposeEE     & 30.5             & 28.3             & 42.6          & 45.8          & 23.6                     & \multicolumn{1}{l|}{22.9} & 29.0          & 27.4                   & 20.8                              & \textbf{19.4} & 33.7           & 39.2          & 16.5          & 55.4          & 30.9          & 27.6          \\
                              & AEC             & \textbf{32.9}    & \textbf{30.2}    & \textbf{46.2} & \textbf{50.1} & \textbf{26.7}            & \textbf{25.2}             & \textbf{31.5} & \textbf{29.8}          & \textbf{22.6}                     & 18.8          & \textbf{37.9}  & \textbf{43.1} & \textbf{17.9} & \textbf{58.6} & 28.8          & 26.7 \\ \hline
\multirow{3}{*}{GPT4o}        & GuidelineEE     & 40.7             & 38.5             & 53.4          & 55.9          & 32.2                     & 33.9                      & 38.9          & 37.8                   & 30.2                              & 29.5          & 44.2           & 50.1          & 19.7          & 59.7          & 36.8          & 35.3          \\
                              & DecomposeEE     & 42.9             & 40.9             & 55.7          & 58.4          & 34.4                     & 35.3                      & 41.2          & 39.9                   & 32.5                              & 31.0          & 47.3           & 52.6          & 21.0          & 62.1          & 37.9          & 36.5          \\
                              & AEC             & \textbf{44.6}    & \textbf{42.8}    & \textbf{58.3} & \textbf{61.8} & \textbf{38.2}            & \textbf{36.8}             & \textbf{43.7} & \textbf{41.9}          & \textbf{34.2}                     & \textbf{32.5} & \textbf{50.8}  & \textbf{56.7} & \textbf{22.5} & \textbf{65.1} & \textbf{39.7} & \textbf{37.8} \\ \hline
\end{tabular}
\caption{Generalization results for ZSEE performance comparing AEC with two major baselines for four other LLMs. \textbf{bold} = best performance. ($\cdot$) = number of distinct event types.}
\label{table:other-llms-zd-results}
\end{table*}

\begin{table}[ht]
\setlength{\tabcolsep}{0.88mm}
\renewcommand{\arraystretch}{1.3}
\centering
\begin{tabular}{lcccccc}
\hline
\multicolumn{7}{c}{\textbf{Llama3-70B}}                                                                                                                  \\ \hline
\multirow{2}{*}{\textbf{Ablation Setting}} & \multicolumn{2}{c}{\textbf{FewEvent}} & \multicolumn{4}{c}{\textbf{ACE}}                         \\ \cline{2-7} 
                                           & TI                   & TC                   & TI            & TC            & AI            & AC            \\ \hline
AEC (full model)                           & \textbf{42.1}        & \textbf{40.5}        & \textbf{57.0} & \textbf{54.6} & \textbf{38.4} & \textbf{34.7} \\
w/o Retrieval Agent                        & 36.5                 & 34.2                 & 49.8          & 47.2          & 33.1          & 30.8          \\
w/o Planning Rationales                    & 38.2                 & 36.0                 & 52.6          & 50.7          & 35.6          & 32.8          \\
w/o Verification Loop                      & 35.0                 & 32.5                 & 47.1          & 44.7          & 30.7          & 28.5          \\
w/o Structural Check                       & 39.8                 & 37.6                 & 54.9          & 52.5          & 37.2          & 33.6          \\ \hline
\multicolumn{7}{c}{\textbf{GPT4o}}                                                                                                                       \\ \hline
\multirow{2}{*}{\textbf{Ablation Setting}} & \multicolumn{2}{c}{\textbf{FewEvent}} & \multicolumn{4}{c}{\textbf{ACE}}                         \\ \cline{2-7} 
                                           & TI                   & TC                   & TI            & TC            & AI            & AC            \\ \hline
AEC (full model)                           & \textbf{44.6}        & \textbf{42.8}        & \textbf{58.3} & \textbf{61.8} & \textbf{38.2} & \textbf{36.8} \\
w/o Retrieval Agent                        & 39.0                 & 36.5                 & 51.6          & 54.2          & 32.6          & 32.0          \\
w/o Planning Rationales                    & 41.3                 & 39.1                 & 54.3          & 57.6          & 35.0          & 33.9          \\
w/o Verification Loop                      & 37.2                 & 34.8                 & 49.0          & 51.3          & 31.1          & 29.5          \\
w/o Structural Check                       & 42.5                 & 40.6                 & 56.7          & 59.9          & 36.4          & 35.2          \\ \hline
\end{tabular}
\caption{Ablation studies on two different LLMs, evaluated on FewEvent and ACE datasets.}
\label{tab:ablation_study}
\end{table}

\paragraph{Base LLMs.} 
We evaluate AEC using a diverse set of instruction‑tuned LLMs from three prominent model families: Llama3‑8B‑Instruct and Llama3‑70B‑Instruct from the Llama3 family~\cite{dubey2024llama}; Qwen2.5‑14B‑Instruct and Qwen2.5‑72B‑Instruct from the Qwen2.5 family~\cite{hui2024qwen2}; and GPT‑3.5‑turbo and GPT‑4o from OpenAI~\cite{achiam2023gpt}. 

\paragraph{Evaluation Metrics.} 
Following prior work~\cite{srivastava2025instruction}, we adopt four standard event extraction metrics:  
(1) \textbf{Trigger Identification (TI)}, which measures the exact match of predicted trigger spans;  
(2) \textbf{Trigger Classification (TC)}, which further requires correct event‑type prediction for each trigger;  
(3) \textbf{Argument Identification (AI)}, which evaluates the accurate extraction of argument spans linked to the predicted triggers; and  
(4) \textbf{Argument Classification (AC)}, the most comprehensive metric, which additionally requires correct role‑type assignment for each argument. We report micro‑averaged F1 scores for all metrics on the constructed test splits.

\paragraph{Implementation Details.} We use TextEE~\cite{huang2024textee} for our benchmarking, datasets. Specifically, AEC is implemented on top of LLM backbones without performing any additional fine‑tuning. We primarily use the instruction‑tuned LLaMA3‑8B and LLaMA3‑70B models as the backbone LLMs. These models power the different agents in the AEC framework and operate entirely in a zero‑shot prompting setting. Each agent is prompted with natural language instructions, and interagent communication is achieved through structured outputs rather than parameter updates (no task‑specific training of the LLMs is performed). For robust evaluation, we report results averaged over three independent runs and set both the number of exemplars and inner‑loop iterations to $k=t=3$. All open‑source models are executed locally on NVIDIA RTX A800 machines equipped with 4 GPUs.

\subsection{Results and Analysis}

\subsubsection{Main Results}
Table~\ref{table:main_results} summarises the primary results comparing AEC with all baseline methods using two variants of Llama3. Across all benchmarks, AEC consistently achieves the best overall performance, substantially outperforming competitive baselines. On ACE 2005 with Llama3‑8B, AEC yields gains of +7.8\% and +6.0\% in TI and TC, respectively, over ChatIE, while also achieving superior results on argument extraction metrics. With Llama3‑70B as the backbone, these improvements become even more pronounced, demonstrating AEC’s strong generalisation ability. 

Although simpler baselines such as DirectEE and GuidelineEE perform reasonably well on smaller or less complex datasets, more structured methods show advantages on datasets with richer schemas. Nevertheless, AEC consistently achieves the best results, particularly on datasets with complex event schemas and diverse event types. These findings confirm that AEC’s collaborative multi‑agent design, coupled with rigorous schema‑driven verification, substantially enhances zero‑shot event extraction performance.

\begin{table*}[ht]

    \centering
    \small
    \begin{tabular}{p{3.5cm}|p{3cm}|p{3cm}|p{3cm}|p{3cm}}
        \toprule
        \textbf{Sentence} & \textbf{Best Baseline} & \textbf{Planning Agent} & \textbf{Coding Agent} & \textbf{Verification Agent} \\
        & \textbf{Prediction} & \textbf{Prediction} & \textbf{Prediction} & \textbf{Prediction} \\
        \midrule
        The company \textbf{acquired} a startup specializing in AI technology. 
        & [(\textbf{``Transaction", ``startup"}), (\textbf{``Acquisition", ``technology"})] 
        & [(``Acquisition", ``acquired")] 
        & [(``Acquisition", ``acquired"), (``Transaction", ``startup")] 
        & [(``Acquisition", ``acquired")] \\
        \midrule
        A massive \textbf{earthquake} struck the city on Monday morning. 
        & [(\textbf{``Disaster", "struck"})] 
        & [(``Earthquake", ``earthquake")] 
        & [(``Earthquake", ``earthquake"), (``Location", ``city")] 
        & [(``Earthquake", ``earthquake"), (``Location", ``city")] \\
        \midrule
        The president \textbf{announced} new \textbf{sanctions} against the country after the attack. 
        & [(\textbf{``Attack", ``announced"})] 
        & [(``Announcement", ``announced")] 
        & [(``Announcement", ``announced"), (``Sanction", ``sanctions")] 
        & [(``Announcement", ``announced"), (``Sanction", ``sanctions")] \\
        \bottomrule
    \end{tabular}
    \caption{Qualitative examples comparing AEC components with the best baseline. Gold triggers and incorrect predictions are in bold.}
    \label{tab:qual-study}
\end{table*}

\subsubsection{Generalization across LLMs}
Table~\ref{table:other-llms-zd-results} demonstrates the generalization capability of AEC across four additional LLMs. AEC consistently achieves top performance over all baselines, with notable average improvements of approximately +3–5\% TI, +4–6\% TC, and +2–4\% in argument metrics compared to the strongest baseline, DecomposeEE. We also observe clear parameter scaling effects: GPT4o achieves the highest overall performance, followed closely by Qwen2.5-72B, underscoring the enhanced reasoning capabilities afforded by larger model sizes under the AEC framework.

\begin{table}[ht]
\centering
\begin{tabular}{c|c|cc|ccll}
\hline
\multirow{2}{*}{$\textbf{\textit{k}}$} & \multirow{2}{*}{$t$} & \multicolumn{2}{c|}{FewEvent} & \multicolumn{4}{c}{ACE}   \\
                     &                      & TI            & TC            & TI   & TC   & AI   & AC   \\ \hline
1                    & 1                    & 40.5          & 39.2          & 52.4 & 52.4 & 31.7 & 29.1 \\
1                    & 3                    & 42.7          & 41.4          & 54.6 & 56.8 & 36.8 & 33.7 \\
3                    & 1                    & 44.1          & 42.0          & 56.2 & 58.7 & 36.3 & 33.4 \\
3                    & 3                    & 44.6          & 42.8          & 58.3 & 61.8 & 38.2 & 36.8 \\
5                    & 3                    & 45.2          & 42.3          & 58.4 & 61.5 & 38.9 & 35.8 \\
5                    & 5                    & 45.1          & 42.7          & 58.3 & 62.3 & 39.1 & 36.2 \\ \hline
\end{tabular}
\caption{Impact of the number of hypotheses $\textbf{\textit{k}}$ and patch attempts $t$ on GPT4o. Performance improves with larger $\textbf{\textit{k}}$ and $t$ but saturates beyond $\textbf{\textit{k}}=3$ and $t=3$.}
\label{tab:impact}
\end{table}

\subsubsection{Ablation Study}
Table~\ref{tab:ablation_study} presents the ablation results on LLaMA3-70B and GPT-4o across the FewEvent and ACE datasets. Removing the \textit{Retrieval Agent} leads to substantial declines in trigger identification performance, highlighting the importance of exemplar generation for contextual disambiguation. Excluding \textit{Planning Rationales} also results in performance degradation, confirming that intermediate reasoning plays a crucial role in guiding accurate type and argument decisions. Disabling the \textit{Verification Loop} or the \textit{Structural Check} consistently reduces all evaluation metrics—particularly for argument classification—demonstrating that code-level validation is critical for enforcing schema fidelity. Overall, these results underscore the necessity of both multi-agent reasoning and schema-as-code verification for achieving robust zero-shot structured event extraction.

\subsubsection{Qualitative Study}
Table~\ref{tab:qual-study} presents qualitative examples comparing AEC components with the best baseline. 
We observe three key trends. 
(i) The Planning Agent produces plausible trigger-type hypotheses but may omit essential arguments. 
(ii) The Coding Agent incorporates more structured arguments guided by the event schema, reducing role confusion. 
(iii) The Verification Agent further corrects type errors and removes inconsistent arguments via schema-level checks. In contrast, the baseline often misclassifies triggers or fails to capture critical arguments. These cases illustrate that AEC's multi-agent reasoning and schema-as-code verification jointly improve contextual disambiguation and structural fidelity in zero-shot event extraction.

\subsubsection{Impact of $\textbf{\textit{k}}$ and $t$}
We study how the number of hypotheses $\textbf{\textit{k}}$ and the maximum patch attempts $t$ affect AEC’s performance. Table~\ref{tab:impact} shows results on FewEvent and ACE using GPT‑4o as the backbone. Increasing $\textbf{\textit{k}}$ provides a larger hypothesis pool, while increasing $t$ allows more opportunities for iterative error correction. Both factors lead to performance improvements up to a certain point, beyond which gains saturate or slightly decline due to the introduction of noisy low‑confidence hypotheses and redundant refinement steps.
\begin{figure}[htbp]
	\centerline{\includegraphics[width=0.9\columnwidth, scale=0.9]{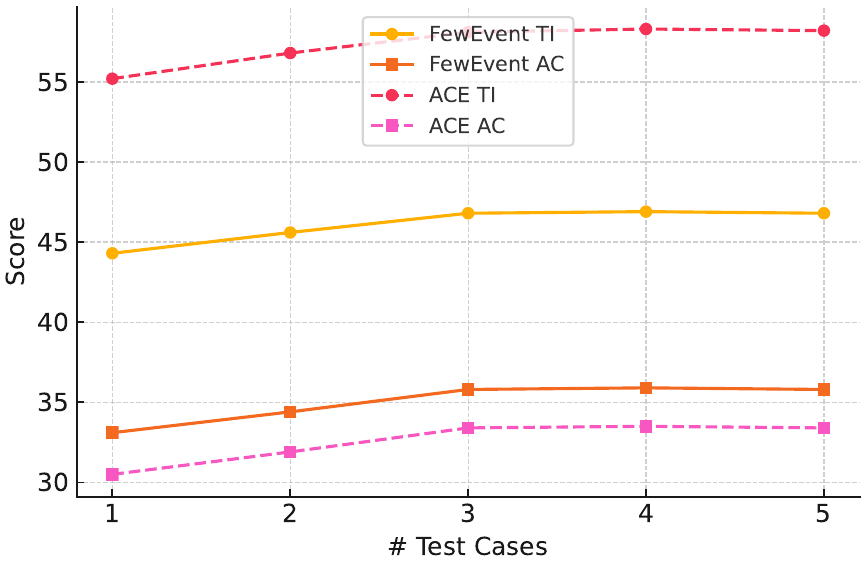}}
	\caption{Impact of the number of test cases in verification.}
	\label{fig:exp1}
\end{figure}

\subsubsection{Impact of Number of Test Cases}
We further examine how the number of test cases in the verification stage influences AEC. Figure ~\ref{fig:exp1} reports results on GPT‑4o. Using more test cases improves both trigger identification and argument classification, as additional checks reduce structural and semantic errors. However, improvements plateau beyond three cases, indicating that further tests add little benefit while increasing computational overhead.

\section{Conclusion}
In this work, we introduced Agent‑Event‑Coder (AEC), a multi‑agent framework that reframes ZSEE as a structured, iterative code‑generation task. By decomposing extraction into Retrieval, Planning, Coding, and Verification agents and representing event schemas as executable classes, AEC enables systematic disambiguation, schema enforcement, and error correction. Experiments on five benchmarks and six LLM backbones show that AEC consistently outperforms strong zero‑shot baselines, especially on complex schemas. Our results demonstrate that combining multi‑agent reasoning with schema‑as‑code verification provides a robust paradigm for zero‑shot structured prediction with LLMs.


\bibliography{aaai2026}

\clearpage
\appendix
\section{Appendix A: Dataset Statistics}

Our experimental setup is a pure zero-shot setup where we do not use any training data.
We provide statistics about the evaluation splits of the different datasets in Table~\ref{tab:data-statistics}.
We follow TextEE \cite{huang2024textee} for the evaluation setup and consider a uniform random split of 250 test samples from each dataset to avoid any train-test split bias.
Since CASIE is a smaller dataset, we only use 50 test samples for this dataset.
The table highlights the domain diversity of the datasets covering common domains like news and general, while also focusing on technical domains like biomedical and epidemiology.
The datasets also show variation in the density, with ACE, FewEvent, and SPEED being sparse with ~1 event mention/sentence.
On the other hand, CASIE, and GENIA are denser with 2.5-10 event mentions/passage.
Finally, we also show the variation in token length, with ACE being the lowest with 13 average tokens, while GENIA and CASIE are longer with 250-280 average tokens per document.
\begin{table}[t]
    \centering
    \small
    \setlength{\tabcolsep}{3.5pt}
    \begin{tabular}{llccc}
        \toprule
        \textbf{Dataset} & \textbf{Domain} & \textbf{\# Doc} & \textbf{\# Event} & \textbf{Avg. Doc} \\
        & & & \textbf{Mentions} & \textbf{Length} \\
        \midrule
        FewEvent & General & 250 & 250 & 30.5 \\
        ACE & News & 250 & 71 & 13.2 \\
        GENIA & Biomedical & 250 & 2472 & 251.3 \\
        SPEED & Epidemiology & 250 & 258 & 32.4 \\
        CASIE & Cybersecurity & 50 & 291 & 283.1 \\
        \bottomrule
    \end{tabular}
    \caption{Data Statistics of the various EE datasets used in our experimental setup.}
    \label{tab:data-statistics}
\end{table}

\begin{table}[h]
    \centering
    \small
    \begin{tabular}{lc}
        \toprule
        \textbf{Dataset} & \textbf{\% Multi-word Triggers} \\
        \midrule
        FewEvent & 3\% \\
        ACE & 2.8\% \\
        GENIA & 8.5\% \\
        SPEED & 0\% \\
        CASIE & 54.6\% \\
        \bottomrule
    \end{tabular}
    \caption{Measuring the percentage of multi-word triggers across the different EE datasets.}
    \label{tab:multi-trigger-analysis}
\end{table}

Different datasets have varied annotation instructions and definitions for the trigger spans.
Some datasets are strictly adhering to only single-word triggers (e.g., SPEED), while others are largely loose and support multi-word triggers (e.g., CASIE).
We also provide a small study of measuring multi-word triggers in Table~\ref{tab:multi-trigger-analysis}, highlighting this disparity across datasets.

\section{Appendix B: Prompt Design}

The ACE framework relies on carefully crafted
prompts to steer a LLM toward producing useful
outputs at each stage of the pipeline.  This appendix describes the
prompt design for each agent and provides concrete examples.  All prompts
are formatted for a chat‑based model with explicit system and user
messages.

\subsection{General Design Principles}

When designing prompts, we follow several guiding principles:

\begin{itemize}
  \item \textbf{Role specification}: Each prompt starts with a clear system
    message that sets the role of the LLM (e.g., ``You are an event
    extraction assistant.'').  This primes the model to behave
    appropriately.
  \item \textbf{Context provision}: The user message includes all
    necessary context, such as event type definitions, datasets,
    exemplar sentences, or candidate hypotheses.  Clear delimiting
    markers (``Event definitions:'', ``Text:'') help the model
    distinguish between different input components.
  \item \textbf{Expected output specification}: We explicitly state
    the desired output format (e.g., ``Return a JSON array of
    \{`trigger': str, `event\_type': str\} objects.'').  This reduces
    ambiguity and helps with automatic parsing.
  \item \textbf{Non‑overlapping scopes}: Each agent's prompt focuses
    solely on its task.  For example, the planning agent extracts
    trigger–event pairs, while the verification agent only checks
    structural and semantic constraints.
\end{itemize}

\subsection{Retrieval Agent Prompts}

The retrieval agent synthesises exemplar sentences that illustrate how an event might be expressed in natural language.  Because retrieving exemplars is optional in some configurations, prompts for this agent are used sparingly.  A typical prompt asks the model to generate a sentence containing a specified event type and its roles.

\paragraph{Design}  The system message sets the role as a ``helpful example generator''.  The user message lists the event type and its argument roles with informal descriptions.  The model is instructed to write a single, fluent sentence that mentions the trigger and all arguments.

\begin{lstlisting}
System: You are a helpful example generator for event extraction.
User: Event type: Databreach
Roles: tool, number-of-data, victim, time, place
Write one English sentence that contains a clear mention of the
Databreach trigger and populates all roles.
\end{lstlisting}

The LLM might respond with a sentence such as ``Hackers used malware to breach the company's database last Tuesday, stealing 10,000 customer records from its New York office.''

\subsection{Planning Agent Prompts}

The planning agent identifies candidate event triggers and their types
within a given text.  It leverages event definitions to constrain the
possible event types and asks the LLM to output a structured list of
matches.

\paragraph{Design}  The system message instructs the model to act as
an ``event extraction assistant''.  The user message contains a block
of Python dataclass definitions representing all event types for the
target dataset and the input text.  A clear instruction directs the
model to return a JSON array of objects with \texttt{trigger} and
\texttt{event\_type} fields.

\begin{lstlisting}
System: You are an assistant for event extraction. Given a piece
of text and definitions of event types (as Python dataclasses),
produce a JSON array of objects where each object has keys 
'trigger' and 'event_type'.

User: Event definitions:
@dataclass
class Databreach:
    mention: str
    tool: List
    number-of-data: List
    victim: List
    time: List
    place: List

@dataclass
class Ransom:
    mention: str
    tool: List
    damage-amount: List
    victim: List
    time: List
    price: List
    place: List

Text:
Hackers demanded a million dollar ransom after infiltrating 
the bank's servers on Friday.

Return a JSON array of {'trigger': str, 'event_type': str} objects.
\end{lstlisting}





The expected answer could be:

\begin{lstlisting}
[
  {"trigger": "demanded", "event_type": "Ransom"},
  {"trigger": "infiltrating", "event_type": "Databreach"}
]
\end{lstlisting}

\subsection{Coding Agent Prompts}

The coding agent converts a selected trigger hypothesis into a concrete
event object, filling argument roles as needed.  In a simple baseline
implementation, argument lists may be left empty, but more advanced
versions can extract values from the text with an additional prompt.

\paragraph{Design}  The system message introduces the role as a
``coding agent'' responsible for generating Python code that
instantiates an event object.  The user message provides the event
definition, the trigger, and the original text.  The model is asked
either to return a JSON representation of the event object with
populated arguments or to output Python code that constructs it.

\begin{lstlisting}
System: You are a coding agent that creates event objects based on
a trigger hypothesis.  Given an event definition, a trigger word,
and the original text, output Python code that imports EventObject
and instantiates it with the appropriate arguments.
User: Event definition:
@dataclass
class Databreach:
    mention: str
    tool: List
    number-of-data: List
    victim: List
    time: List
    place: List

Trigger: "breach"
Text: "The attacker executed a data breach using phishing emails on
Wednesday, compromising 5,000 records at the hospital."
\end{lstlisting}

An appropriate response would extract values for roles (e.g., tool =
``phishing emails'', number‑of‑data = ``5,000'', victim = ``hospital'',
time = ``Wednesday'') and produce code to construct an
\texttt{EventObject} with these arguments.

\subsection{Verification Agent Prompts}

The verification agent checks whether a candidate event object adheres
to the schema and is consistent with the source text.  While our
implementation performs programmatic checks, one can also employ the
LLM to reason about errors and suggest corrections.

\paragraph{Design}  The system prompt positions the model as a
``verifier'' that inspects an event object and returns feedback.
The user supplies the original text, the event definition, and the
candidate event object (either as code or JSON).  The model is asked to
judge three aspects: (1) Does the trigger appear in the text? (2) Are
all argument values of the correct type? (3) Do the roles match the
definition?

\begin{lstlisting}
System: You are a verifier for event extraction.  Examine the
following event object in the context of the original text and
definition.  Identify any semantic, type, or structural errors.
User: Text: "On Tuesday the company patched a vulnerability in its
web server."  Event definition:
@dataclass
class PatchVulnerability:
    mention: str
    patch: List
    cve: List
    time: List
    vulnerable_system: List

Candidate event object:
{
  "event_type": "PatchVulnerability",
  "trigger": "patched",
  "arguments": {
    "patch": ["security update"],
    "cve": ["CVE-2021-1234"],
    "time": ["Tuesday"],
    "vulnerable_system": [1234]
  }
}

Return a list of error messages, or an empty list if the event is
valid.
\end{lstlisting}

In this example the model should flag that the value ``1234'' for
\texttt{vulnerable\_system} is not of type \texttt{str} (a type error).

\section*{Summary}

By defining clear roles, supplying precise context, and stating the
desired output format, the AEC pipeline prompts enable large language
models to collaborate effectively across retrieval, planning, coding,
and verification stages.  These examples can serve as a template for
designing prompts tailored to new datasets and event ontologies.

\end{document}